\documentclass[10pt,twocolumn,letterpaper]{article}

\usepackage{bm}
\usepackage{multirow}
\usepackage{amsfonts}
\usepackage{amsmath}
\usepackage[table]{xcolor}
\usepackage[pagenumbers]{cvpr} % To force page numbers, e.g. for an arXiv version
\definecolor{cvprblue}{rgb}{0.21,0.49,0.74}

\usepackage{multirow}
\usepackage[pagebackref,breaklinks,colorlinks,allcolors=cvprblue]
{hyperref}

\def\paperID{} %14723 *** Enter the Paper ID here
\def\confName{CVPR}
\def\confYear{2026}

\title{From Patches to Evidence Balls: Class-Conditioned Evidence Retrieval for Few-Shot Whole Slide Image Classification}
\author{
Di Zhang$^{1}$ \quad Li Zhang$^{2}$\footnotemark[1]   \quad Jiashuai Liu$^{1}$ \quad Junbo Lu$^{1}$ \quad Zhi Zeng$^{1}$ \quad Jiusong Ge$^{1}$ \\
Chunze Yang$^{1}$ \quad
 Yi Niu$^{1}$ \quad Jian Chen$^{3}$ \quad Kai He$^{4}$ \quad Zeyu Gao$^{3}$\thanks{Co-corresponding authors. $<$ zg323@cam.ac.uk$>$}\quad Chen Li$^{1}$ \\
{
$^{1}$XJTU \quad
$^{2}$JUFE \quad
$^{3}$University of Cambridge \quad
$^{4}$NUS
}
}

\begin{document}
\maketitle
\begin{abstract}
Whole slide image (WSI) classification is an evidence-driven task, where diagnostic cues are often sparse, spatially organized, and class-dependent. Existing multiple instance learning (MIL) and vision-language methods aggregate a large pool of patch features into a single global slide representation. Under few-shot supervision, limited slide-level labels make it difficult to learn a reliable aggregation mechanism that organizes sparse local cues into compact and coherent diagnostic evidence. Moreover, a shared slide representation compresses evidence supporting a candidate class and its alternatives into the same feature, limiting class-specific reasoning and interpretability. To address these issues, we propose \textbf{EviBall}, a class-conditioned evidence retrieval framework for few-shot WSI classification. EviBall organizes local patches into \textit{Evidence Balls} through semantic-spatial assignment and center refinement, yielding compact and spatially coherent evidence units under weak supervision. It then uses task-specific class queries, including language-guided queries for morphology-oriented tasks and molecular-guided queries for molecular endpoint prediction, to retrieve supporting evidence balls and produce class-conditioned evidence representations for direct class-wise prediction. By introducing structured evidence units and task-relevant semantic guidance, EviBall reduces the reliance on learning an unconstrained global aggregation mechanism from scarce slide-level labels. It therefore reformulates few-shot WSI classification as structured evidence retrieval and competition among candidate classes. Extensive experiments across four morphology-oriented and molecular endpoint WSI tasks demonstrate that EviBall consistently outperforms conventional and vision-language MIL baselines under diverse few-shot settings, while providing spatially localized and class-specific evidence for each prediction. 
\end{abstract}    
\section{Introduction}
\label{sec:intro}
Histopathological examination plays a central role in disease diagnosis, tumor grading, biomarker assessment, and prognosis prediction. 
With the increasing digitization of glass slides into whole-slide images (WSIs), computational pathology has enabled data-driven models to assist pathologists in large-scale and objective tissue analysis \cite{song2023artificial}. 
However, WSI classification \cite{niu2025learning} is fundamentally different from conventional natural image recognition. 
Unlike natural images, where a prediction can often be made from a holistic object-level representation, a WSI is classified based on sparse, spatially organized, and class-dependent histological evidence distributed across a gigapixel-scale tissue landscape \cite{li2026turning}. 
Such evidence may occupy only a small fraction of the entire slide and is often embedded in highly heterogeneous tissue contexts \cite{niu2026age}.

%Histopathological examination plays a central role in disease diagnosis, tumor grading, molecular subtype assessment, and prognosis prediction. With the increasing digitization of glass slides into whole slide images (WSIs), computational pathology has enabled data-driven models to assist pathologists in large-scale and objective tissue analysis. However, WSI classification is fundamentally different from conventional natural image recognition. Unlike natural images, where a prediction can often be made from a holistic object-level representation, a WSI is usually classified based on sparse, spatially organized, and class-dependent histological evidence distributed across a gigapixel-scale tissue landscape. Such evidence may occupy only a small fraction of the entire slide and is often embedded in highly heterogeneous tissue contexts. Therefore, WSI classification should be viewed as a process of constructing clinically meaningful evidence units from heterogeneous patch observations and decoding class-conditioned evidence that supports each candidate diagnosis.

\begin{figure}[t]
\centering
\includegraphics[width=\columnwidth]{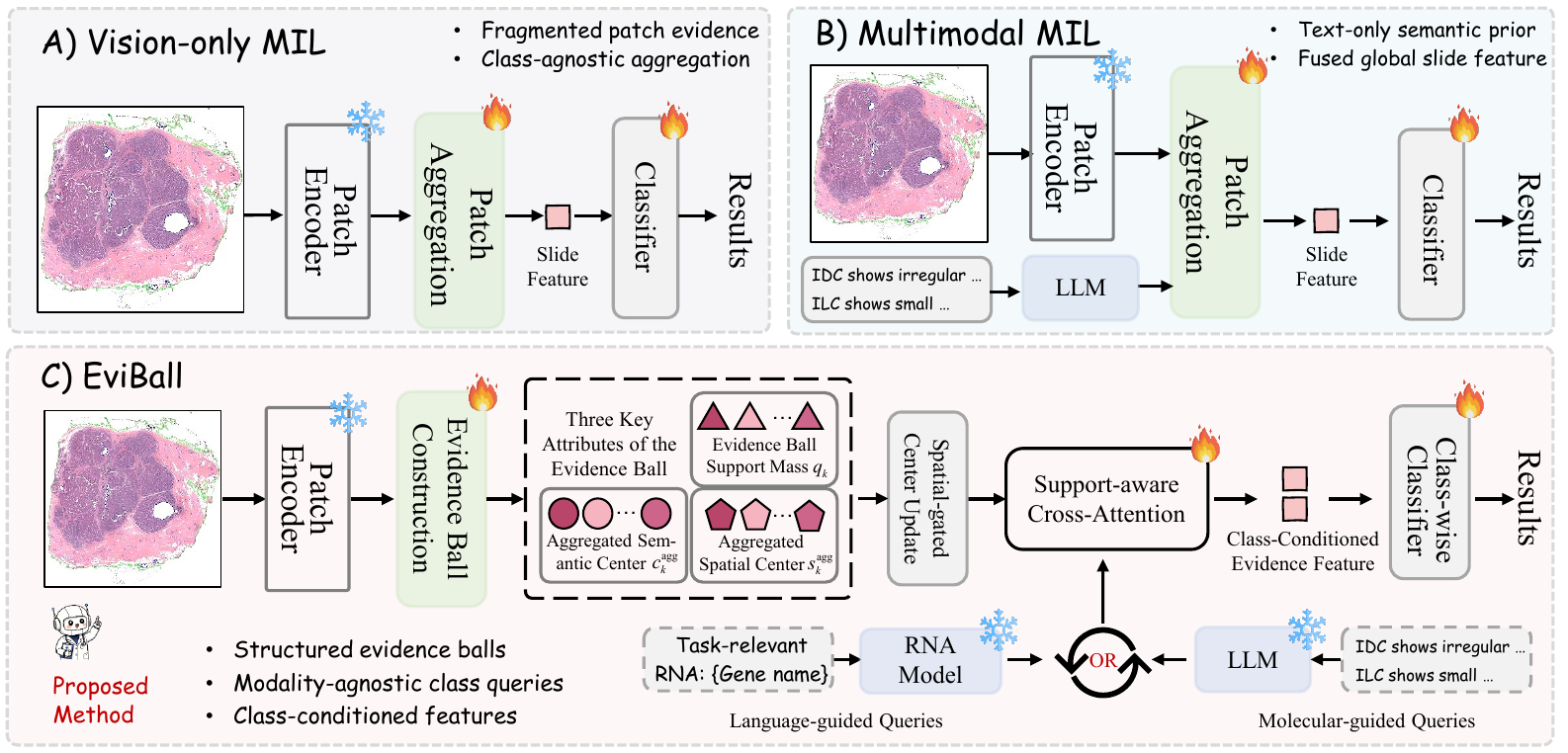} % Reduce the figure size so that it is slightly narrower than the column.
\caption{Motivation of EviBall. Vision-only MIL relies on fragmented patch-level evidence, while LLM-based multimodal MIL often uses text-limited semantic guidance and global feature fusion. EviBall constructs structured evidence balls and performs class-conditioned evidence retrieval with task-relevant semantic queries.}
\label{motivation}
\end{figure}

This evidence-centric nature~\cite{yan2026llm} makes few-shot WSI classification particularly challenging.
To process gigapixel WSIs using only slide-level supervision, existing methods commonly adopt multiple instance learning (MIL), representing each slide as a bag of patch features and aggregating them into a slide-level representation for prediction.
However, in many clinically relevant tasks~\cite{zhang2025stadis,li2025similarity}, reliable slide-level labels are costly and time-consuming to obtain, as they often require molecular assays, expert pathological review, or long-term clinical follow-up.
As a result, only a handful of labeled slides may be available for each class.
Under such limited supervision, the model must identify class-discriminative evidence from thousands of heterogeneous patches, although only a small fraction may be relevant to the target endpoint.
This makes it difficult to learn a reliable aggregation mechanism that preserves sparse diagnostic evidence.

%However, existing WSI classification paradigms do not adequately support both structured evidence modeling and task-relevant semantic guidance. As illustrated in Fig.~\ref{motivation}, vision-only MIL methods \cite{abmil} usually represent a WSI as a bag of patch instances and directly aggregate them into a global slide-level feature.  Although this paradigm avoids dense region annotation, it remains patch-centric: individual tiles serve as the basic modeling units, while their local semantic and spatial relationships are not explicitly organized. Consequently, the evidence used for prediction can be fragmented, unstable, and class-agnostic, especially when only a few labeled slides are available.  Recent LLM-based multimodal MIL methods \cite{li2026universal} introduce textual semantics to provide class-level guidance, but they are often text-limited and fusion-centric.  Textual descriptions may provide useful priors for morphology-oriented tasks, yet they can be indirect for molecular endpoints defined by RNA expression, mutation status, or biomarker states.  Moreover, most multimodal methods still fuse visual and textual information into a global slide representation, where class-specific supporting evidence and competing evidence may remain entangled.  These limitations suggest that few-shot WSI classification requires not only more informative evidence units, but also task-relevant class semantics that can decode evidence in a class-conditioned manner.

However, existing MIL paradigms are not explicitly designed to organize such sparse local cues into reliable and class-specific evidence.
As illustrated in Fig.~\ref{motivation}, vision-only MIL methods~\cite{clam} typically aggregate fragmented patch evidence into a class-agnostic global slide representation \cite{liu2024correlation}.
Moreover, the spatial organization and support strength of local evidence are often not explicitly modeled, causing diagnostically relevant cues to remain fragmented or be diluted during aggregation.
Recent multimodal MIL methods~\cite{li2026universal} introduce semantic guidance, but commonly rely on text-only priors and fuse semantic information into a shared slide feature.
Such language guidance may be informative for morphology-oriented tasks but indirect for molecular endpoints, while global fusion can entangle the evidence supporting candidate classes.
These limitations call for organizing local cues into coherent evidence units, incorporating task-relevant semantics, and preserving class-specific evidence under limited supervision.

Motivated by these observations, we propose EviBall, a class-conditioned evidence retrieval framework that recasts few-shot WSI classification as structured evidence reasoning rather than global slide representation learning. EviBall first organizes fragmented patch instances into \textit{Evidence Balls}, compact evidence units that jointly encode semantic content, spatial organization, and evidence support mass. These structured units are then queried using task-relevant class semantics, derived from language descriptions for morphology-oriented tasks or RNA representations for molecular endpoint prediction. Through support-aware cross-attention, each class query selectively retrieves its supporting evidence balls to form a distinct class-conditioned representation for prediction. To further reduce redundant evidence, we introduce a same-region diversity regularizer that encourages co-located evidence balls to capture complementary local patterns.
We evaluate EviBall on multiple few-shot WSI classification tasks under 1-, 2-, 4-, and 8-shot settings, covering both morphology-oriented classification and molecular endpoint prediction. 
Compared with representative vision-only MIL methods and LLM-based multimodal MIL baselines, EviBall achieves strong and generally superior performance across tasks and shot numbers, demonstrating the effectiveness of structured evidence construction and class-conditioned evidence retrieval. 
Notably, on molecular endpoint prediction, the molecular-guided variant further improves or complements the LLM-guided variant, suggesting that task-relevant molecular semantics can provide more appropriate class guidance than text-only priors. 

Our main contributions are summarized as follows:
%To realize this formulation, we propose EviBall, a class-conditioned evidence decoding framework that constructs Evidence Balls from patch-level observations and uses class textual queries to retrieve supporting evidence for each candidate class. EviBall first moves beyond isolated patch instances by constructing Evidence Balls as intermediate tissue units. Each Evidence Ball is designed to summarize a local group of patches into a semantically discriminative and spatially coherent representation, thereby serving as a more stable evidence unit than individual patches. On top of these structured tissue units, EviBall introduces a text-guided evidence decoder. Class-specific textual prompts are encoded as semantic queries, and each query attends to the Evidence Balls to retrieve the tissue evidence most relevant to its corresponding class. The retrieved class-wise evidence representations are then directly used to compute class logits, rather than being collapsed into a shared slide embedding. In this way, EviBall reformulates few-shot WSI classification as a process of structured evidence construction, class-conditioned evidence retrieval, and evidence competition among candidate classes.

%Patch is the observation unit, slide embedding is the prediction unit, but neither is necessarily the right evidence unit. EviBall introduces Evidence Balls as intermediate tissue evidence units and uses language as a class-conditioned evidence decoder.

%In summary, our contributions are as follows:
\begin{itemize}
    \item We propose EviBall, an evidence-centric framework for few-shot WSI classification that introduces evidence balls to shift the modeling focus from isolated patch instances to compact, spatially coherent evidence units.

    \item We introduce task-adaptive semantic querying that extends multimodal MIL beyond language-only guidance. EviBall employs LLM-derived morphological semantics for morphology-oriented tasks and molecular-derived semantics for molecular endpoint prediction.
    
    \item We develop a support-aware, class-conditioned evidence decoder, in which each class query selectively retrieves semantically relevant evidence balls to form a dedicated evidence representation for class-wise prediction.

    \item Extensive experiments on multiple few-shot WSI classification tasks demonstrate the effectiveness of EviBall against representative vision-only and multimodal MIL baselines, with the molecular-guided variant showing additional benefits for molecular endpoint prediction.
\end{itemize}

%-------------------------------------------------------------------------

\section{Related Works}
\label{sec:formatting}

\paragraph{MIL and Multimodal Learning for WSI Classification.}
Due to the gigapixel scale of WSIs and the scarcity of dense annotations, multiple instance learning has become a dominant weakly supervised paradigm for WSI classification~\cite{yao2020whole}. Existing methods improve slide-level modeling through attention pooling~\cite{clam}, instance selection~\cite{fu2010milis}, dual-stream learning~\cite{li2021dual}, transformer-based contextualization~\cite{transmil}, and few-shot feature adaptation or regularization~\cite{ wong2026few}. Multimodal approaches further incorporate semantic priors from class names, textual descriptions, or vision-language models~\cite{gao2025alpaca,shi2024vila,huang2023visual}. However, most methods remain patch-centric and aggregate or fuse patch features into a global slide representation, without explicitly organizing them into structured evidence units for class-wise retrieval.
\begin{figure*}[t]
\centering
\includegraphics[width=\textwidth]{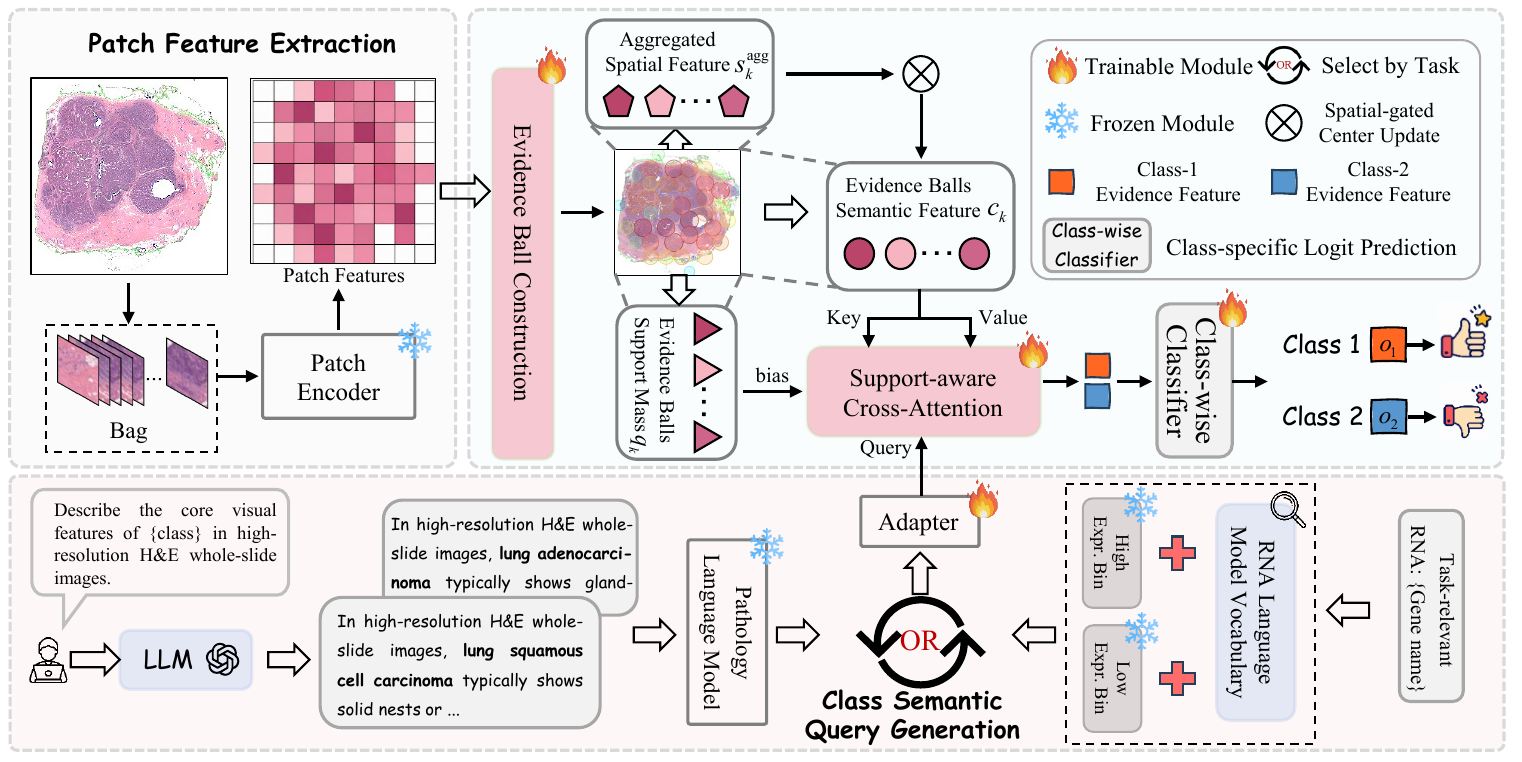} % Reduce the figure size so that it is slightly narrower than the column.
\caption{Framework of EviBall. Given a WSI, patch features are first extracted by a frozen patch encoder and organized into structured evidence balls with semantic, spatial, and support mass attributes. Task-relevant class semantic queries are then generated from either language or RNA semantics. Through support-aware cross-attention, each class query retrieves its supporting evidence balls to form a class-conditioned representation, which is finally used for class-wise slide-level prediction. }
\label{framework}
\end{figure*}
\paragraph{Evidence-Centric Representation in WSI Analysis.}
Recent studies have moved toward evidence-centric modeling by identifying discriminative patches~\cite{transmil}, learning prototypes~\cite{shi2024vila}, performing multi-agent diagnostic reasoning \cite{weishaupt2025evidence} or constructing graphs~\cite{zheng2022graph}. Nevertheless, these evidence units are typically represented primarily by semantic features. EviBall constructs evidence balls characterized by semantic centers, spatial centers, and relative support mass, enabling class-conditioned reasoning over explicit evidence units. A detailed comparison is provided in the Appendix.

\section{Methodology}
\label{sec:method}
   %3.1 Problem Formulation
   \subsection{Problem Formulation}
A WSI is a gigapixel-scale image that cannot be directly processed in the same way as conventional natural images. A common practice in computational pathology is therefore to divide each WSI into a set of smaller patches and extract patch-level features using a pretrained encoder.
Let $\mathcal{D}=\{(X_m,y_m)\}_{m=1}^{M}$ denote a weakly supervised WSI classification dataset, where $X_m$ is the $m$-th whole slide image and $y_m\in\{1,\dots,C\}$ is its slide-level label over $C$ classes. Following common WSI pipelines, each slide is divided into a set of non-overlapping patches and encoded by a pretrained patch encoder. Thus, a WSI is represented as a bag of patch-level instances: $X_m=\{(h_i,p_i)\}_{i=1}^{N},$ where $h_i\in \mathbb{R}^{d}$ denotes the feature of the $i$-th patch, $p_i\in\mathbb{R}^{2}$ denotes its spatial coordinate, and $N$ is the number of patches in the slide. During training, only the slide-level label $y_m$ is available, while patch-level diagnostic annotations are not provided.

%Conventional MIL methods usually formulate WSI classification as global slide representation learning. Given patch features $\{h_i\}_{i=1}^{N}$, they aggregate all patch instances into a slide embedding:
%\begin{equation} s = \mathrm{MIL}(\{h_i\}_{i=1}^{N}), \quad o = f(s),\end{equation}
%where $s$ is used to predict the logits of all candidate classes. Although effective, this formulation makes different class predictions from the same global representation, where class-specific evidence may be entangled with competing cues and non-diagnostic context.

   %3.2 Overview of EviBall

To enable few-shot WSI classification with structured and task-relevant evidence, we propose EviBall, as illustrated in Fig.~\ref{framework}. EviBall comprises three components: a frozen CONCH v1.5 encoder~\cite{conch} followed by evidence ball construction to form compact evidence units with semantic, spatial, and support mass information, task-relevant class semantic queries derived from language or RNA queries, and support-aware cross-attention that retrieves class-specific evidence for class-wise prediction.

   %3.3 Evidence Ball Construction
\subsection{Evidence Ball Construction}
\label{sec:evidence_ball_construction}

\begin{figure}[t]
\centering
\includegraphics[width=0.9\columnwidth]{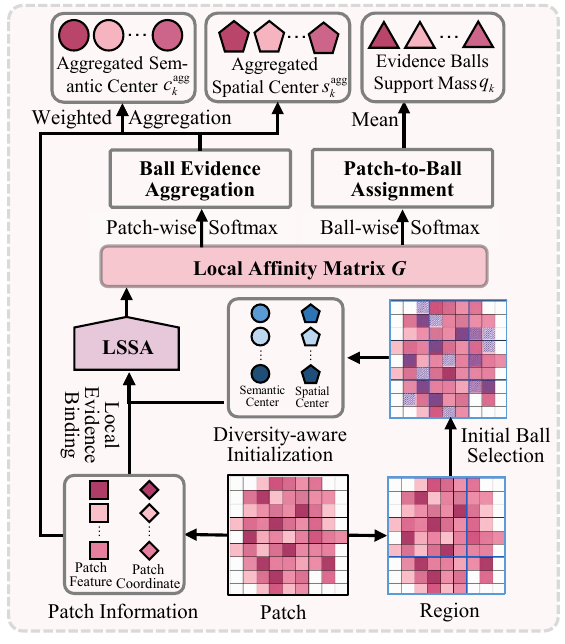} % Reduce the figure size so that it is slightly narrower than the column.
\caption{Overview of Evidence Ball Construction. }
\label{module}
\end{figure}

Evidence ball construction aims to move beyond isolated patch instances by grouping patch-level observations into compact structured evidence ball units. Given a WSI bag $X_m$, EviBall constructs a set of evidence balls
\begin{equation}
\mathcal{B}_m=\{B_k\}_{k=1}^{K}
=\{(c_k,s_k,q_k)\}_{k=1}^{K},
\end{equation}
where $c_k$, $s_k$, and $q_k$ denote the semantic center, spatial center, and relative support mass of the $k$-th evidence ball, respectively.
The construction consists of diversity-aware initialization, local patch-to-ball assignment, and evidence ball refinement, as illustrated in Fig.~\ref{module}.

\paragraph{Diversity-aware initialization.}
The patch features are first arranged into a spatial feature grid according to their coordinates and then partitioned into non-overlapping local regions. 
Every $M \times M$ neighboring patches are treated as one local region, where $M$ denotes the region size.  Within each valid region $\mathcal{R}_r$, the least similar patch pair is selected to initialize two semantically diverse evidence balls:
\begin{equation}
(i_r,j_r)=
\arg\min_{i,j\in\mathcal{R}_r,\;i\neq j}
\mathrm{sim}(h_i,h_j).
\label{init}
\end{equation}
The two selected seed patches independently initialize two candidate evidence balls:
$B_{i_r}^{\mathrm{init}}=(c_{i_r},s_{i_r})$, and $B_{j_r}^{\mathrm{init}}=(c_{j_r},s_{j_r})$.
Collecting the initialized balls from all valid local regions yields
\begin{equation}
\mathcal{B}_m^{\mathrm{init}}
=
\bigcup_{\mathcal{R}_r}
\{B_{i_r}^{\mathrm{init}},B_{j_r}^{\mathrm{init}}\}
=
\{(c_k^{\mathrm{init}},s_k^{\mathrm{init}})\}_{k=1}^{K}.
\end{equation}
At this initialization stage, each candidate ball only contains an initial semantic center and spatial center; its relative support mass is not computed until the subsequent soft assignment and refinement steps.
By initializing two semantically diverse balls within each local region, EviBall preserves heterogeneous patch patterns as separate candidate evidence.

\paragraph{Local Evidence Binding.}
To transform seed-level candidates into evidence balls, EviBall aggregates semantically consistent patches around each initialized seed within a local spatial neighborhood. 
Specifically, for each initialized ball $B^{\mathrm{init}}_k$, we define a neighboring patch set $\mathcal{N}_k$ consisting of patches from the same or adjacent local regions. 
The subsequent local evidence binding is performed only within $\mathcal{N}_k$, which encourages each ball to capture a coherent local evidence pattern while reducing the noise of global matching.

We introduce a Local Semantic-Spatial Affinity (LSSA) operation to measure the
compatibility between each initialized evidence ball and its neighboring
patches. Specifically, we compute a local affinity matrix
$G=[g_{ki}]\in\mathbb{R}^{K\times N}$, where the $k$-th row corresponds to the
$k$-th evidence ball and the $i$-th column corresponds to the $i$-th patch.
The LSSA score is defined as
\begin{equation}
g_{ki}=
\begin{cases}
\dfrac{(W_Q c_k^{\mathrm{init}})^\top(W_K h_i)}{\sqrt{d}}
-\lambda\dfrac{\|s_k^{\mathrm{init}}-p_i\|_2}{M},
& i\in\mathcal{N}_k,\\
-\infty, & i\notin\mathcal{N}_k,
\end{cases}
\label{LSSA}
\end{equation}
where $W_Q$ and $W_K$ are learnable projection matrices, $d$ is the feature
dimension, and $\lambda$ controls the spatial distance penalty.
From the same masked LSSA logits, EviBall applies a ball-wise softmax and a
patch-wise softmax to derive two complementary normalized weights:
\begin{equation}
\alpha_{ki}=\mathrm{Softmax}_{\mathrm{ball}}(G)_{ki},
\qquad
\beta_{ki}=\mathrm{Softmax}_{\mathrm{patch}}(G)_{ki}.
\end{equation}
Here, the ball-wise softmax normalizes over candidate evidence balls for each
patch and produces patch-to-ball assignment weights, whereas the patch-wise
softmax normalizes over neighboring patches for each evidence ball and produces
evidence aggregation weights.

%The two weights provide different views of the local patch-to-ball relation. The patch-wise membership $\alpha_{ki}$ indicates how much patch $i$ belongs to Evidence Ball $B_k$ among the candidate balls covering this patch. 

We estimate the evidence support mass of $B_k$ by its average patch membership $q_k=\frac{1}{N}\sum_{i=1}^{N}\alpha_{ki}.$
In contrast, the ball-wise contribution $\beta_{ki}$ indicates how much patch
$i$ contributes to refining evidence ball $B_k$ among its neighboring patches.
Using these contribution weights, EviBall composes aggregated semantic
and spatial centers:
\begin{equation}
{c}^{\mathrm{agg}}_k =
\sum_{i=1}^{N}\beta_{ki} W_v h_i,
\quad
{s}^{\mathrm{agg}}_k =
\sum_{i=1}^{N}\beta_{ki} p_i.
\end{equation}

\paragraph{Spatial-gated center update.}
To obtain the final semantic center, EviBall further applies a spatial-gated center update.
The spatial displacement between the provisional spatial center and the aggregated spatial center is computed as
$\Delta s_k=\|{s}^{\mathrm{agg}}_k-s_k^{\mathrm{init}}\|_2$.
We then define a spatial gate:
\begin{equation}
\rho_k=
\frac{\Delta s_k}
{\Delta s_k+\frac{1}{K}\sum_{k'=1}^{K}\Delta s_{k'}}.
\end{equation}
The spatial gate controls the refinement of the semantic center by modulating
how much it moves toward the locally composed semantic evidence:
\begin{equation}
c_k =
\left(
1 - \rho_k
\right)
c_k^{\mathrm{init}}
+
\rho_k
{c}^{\mathrm{agg}}_k,
\quad
s_k= \left(
1 - \rho_k
\right) s_k^{\mathrm{init}}  +\rho_k {s}^{\mathrm{agg}}_k.
\end{equation}
The refined evidence ball is therefore represented as $B_k=(c_k,s_k,q_k)$. The spatial gate further stabilizes center refinement by allowing large semantic updates only when the spatial center is consistently shifted by its local evidence.
%In this design, $\alpha_{ki}$ captures patch-to-ball ownership and is used to estimate the evidence quality, while $\beta_{ki}$ captures ball-to-patch contribution and is used to compose the semantic and spatial centers. 

\subsection{Class Semantic Query Generation}

To retrieve evidence in a class-aware manner, EviBall assigns each candidate class a semantic query. Specifically, we instantiate the queries with language-model-derived semantics for morphology-oriented tasks and molecular-model-derived semantics for molecular endpoint prediction.

\paragraph{Language-guided queries.}
%For morphology-oriented tasks, class semantics are described by visual pathological concepts. For each class $c$, we prompt GPT-5.5 with ``Describe the core visual features of {class} in high-resolution H\&E whole-slide images'' to obtain a class-specific textual description $T_c$. The generated description is then encoded by the frozen TITAN text encoder \cite{titan}, and we use its text [CLS] token as the language-guided semantic anchor. 
For morphology-oriented tasks, class-specific pathological descriptions are
generated by an LLM and encoded by the frozen TITAN text encoder \cite{titan}, whose
$\texttt{[CLS]}$ token serves as the semantic query. This provides a morphology-aware semantic representation for each candidate class. For a fair comparison, all VLM-based MIL methods use the same class descriptions and frozen TITAN text encoder.

\paragraph{Molecular-guided queries.}
For molecular endpoint prediction, textual descriptions may provide only indirect priors because the target labels are defined by molecular states. We therefore construct molecular-guided semantic queries using a pretrained transcriptomic foundation model, scGPT~\citep{cui2023scGPT}. For a binary endpoint associated with a marker gene $g$, the negative and positive classes are represented by low- and high-expression bins, respectively. Following the categorical input formulation of scGPT, the molecular query for class $c$ is defined as
\begin{equation}
u_c=\mathrm{GeneEnc}(g)+\mathrm{ValueEnc}(b_c),
\qquad b_c\in \{b^{-},b^{+}\},
\end{equation}
where $b^{-}$ and $b^{+}$ denote the low- and high-expression states.

\paragraph{Query projection.}
Since the semantic queries from different sources may lie in different embedding spaces, we use a source-specific adapter to project them into the evidence ball space:
$e_c=\mathrm{Adapter}(u_c)\in\mathbb{R}^{d}$, where $c\in\{1,\dots,C\}$.
Here, $e_c$ is the final class semantic query used by the evidence retrieval module. This design allows EviBall to use language-guided queries for morphology-oriented tasks and molecular-guided queries for molecular endpoints within a unified class-conditioned evidence retrieval framework.

   %3.4 Class-Conditioned Evidence Decoding
\subsection{Class-Conditioned Evidence Retrieval}
\label{sec:class_conditioned_decoding}

Given the class semantic queries and the constructed evidence balls, EviBall
performs class-conditioned evidence retrieval for slide-level prediction. Unlike
conventional MIL methods that aggregate all evidence into a single global slide
representation, EviBall allows each class query to retrieve its own supporting
evidence balls, reformulating WSI classification as class-wise evidence
retrieval and competition.

\paragraph{Support-aware evidence retrieval.}
Given the class semantic queries $\{e_c\}_{c=1}^{C}$ and the evidence ball set
$\mathcal{B}_m=\{({c}_k,{s}_k,q_k)\}_{k=1}^{K}$ for the $m$-th
WSI, EviBall performs support-aware cross-attention:
\begin{equation}
Z
=
\mathrm{Softmax}
\left(
\frac{QK^{\top}}{\sqrt{d}}
+
\gamma\,U^{\top}
\right)
V,
\label{xatten}
\end{equation}
where $Q$ is obtained from the class semantic queries by a linear transformation,
and $K$ and $V$ are obtained from the semantic centers of evidence balls through learnable linear projections. Here, $\gamma>0$ is a learnable scale constrained to be positive, ensuring that evidence balls with higher support receive larger additive attention biases, and $d$ is the feature dimension, and $U=[u_1,\ldots,u_{K}]^{\top} \in \mathbb{R}^{k\times 1}$ denotes the standardized support scores
$u_k=\frac{\log q_k-\mu}{\sigma},$
where $\mu$ and $\sigma$ are computed over evidence balls in the WSI. 

The support term $\gamma U^{\top}$ provides an additive evidence support bias,
favoring better-supported evidence balls when their semantic relevance is
comparable.
%The quality term $\gamma U^{\top}$ is added to the attention logits before softmax and is broadcast to all class queries. It provides an evidence reliability prior, encouraging higher-quality Evidence Balls to receive larger attention weights when their semantic relevance is comparable. The softmax is normalized over valid Evidence Balls, and the $c$-th row of $Z$, denoted as $z_c$, is the class-conditioned evidence representation for class $c$.

\paragraph{Class-wise prediction.}
The class-conditioned representations are refined and converted into class
logits by class-specific prediction heads:
\begin{equation}
\tilde{Z}=\mathrm{FFN}(\mathrm{LN}(Z)),\quad
p=\mathrm{Softmax}([g_1(\tilde{z}_1),\ldots,g_C(\tilde{z}_C)]),
\end{equation}
where $\tilde{z}_c$ is the $c$-th row of $\tilde{Z}$, $\mathrm{FFN}$ denotes a feed-forward network, and $g_c(\cdot)$ denotes
the prediction head for class $c$. Since each $z_c$ is decoded from the corresponding class query, the final prediction is based on class-specific evidence representations rather than a single global slide embedding.
   
\begin{table*}[t]
\centering
\small
\setlength{\tabcolsep}{3pt}
\begin{tabular}{llcccccccc}
\toprule
\multirow{2}{*}{\rotatebox{90}{Task}} & \multirow{2}{*}{Methods} & \multicolumn{2}{c}{1-shot} & \multicolumn{2}{c}{2-shot} & \multicolumn{2}{c}{4-shot} & \multicolumn{2}{c}{8-shot} \\
\cmidrule(lr){3-4}\cmidrule(lr){5-6}\cmidrule(lr){7-8}\cmidrule(lr){9-10}
& & ACC & AUC or F1 & ACC & AUC or F1 & ACC & AUC or F1 & ACC & AUC or F1 \\
\midrule
\multirow{9}{*}{\rotatebox{90}{RCC-Subtyping}}
& ABMIL & $54.10_{\pm 11.73}$ & $77.40_{\pm 4.90}$ & $51.47_{\pm 15.30}$ & $81.62_{\pm 8.25}$ & $52.79_{\pm 16.43}$ & $78.21_{\pm 7.87}$ & $73.53_{\pm 11.00}$ & $80.98_{\pm 10.68}$ \\
& CLAM & $51.02_{\pm 7.48}$ & $79.54_{\pm 9.00}$ & $39.05_{\pm 6.93}$ & $72.22_{\pm 9.00}$ & $61.48_{\pm 21.91}$ & $76.88_{\pm 8.43}$ & $\mathbf{86.57_{\pm 5.20}}$ & $77.52_{\pm 6.86}$ \\
& TransMIL & $47.75_{\pm 10.39}$ & $81.65_{\pm 3.74}$ & $52.07_{\pm 12.96}$ & $\mathbf{86.81_{\pm 2.83}}$ & $50.75_{\pm 14.90}$ & $84.17_{\pm 6.86}$ & $72.64_{\pm 12.33}$ & $85.08_{\pm 6.48}$ \\
& RRT & $56.42_{\pm 14.14}$ & $\underline{87.13_{\pm 6.86}}$ & $71.44_{\pm 12.49}$ & $\underline{86.20_{\pm 5.29}}$ & $63.40_{\pm 12.25}$ & $85.61_{\pm 6.63}$ & $80.58_{\pm 15.78}$ & $82.74_{\pm 8.25}$ \\
& DSMIL & $49.53_{\pm 14.11}$ & $59.07_{\pm 15.09}$ & $47.06_{\pm 7.62}$ & $75.29_{\pm 5.83}$ & $58.99_{\pm 14.80}$ & $80.45_{\pm 7.75}$ & $80.31_{\pm 12.96}$ & $80.02_{\pm 11.14}$ \\
& ViLaMIL & $\underline{70.85_{\pm 13.27}}$ & $68.86_{\pm 10.95}$ & $\underline{74.89_{\pm 14.07}}$ & $76.19_{\pm 7.21}$ & $\underline{84.44_{\pm 6.08}}$ & $79.98_{\pm 7.94}$ & $83.53_{\pm 6.78}$ & $78.68_{\pm 10.05}$ \\
& FOCUS & $59.94_{\pm 11.87}$ & $86.61_{\pm 6.40}$ & $64.92_{\pm 7.00}$ & $83.57_{\pm 7.07}$ & $50.92_{\pm 12.53}$ & $84.75_{\pm 7.81}$ & $65.97_{\pm 16.46}$ & $79.76_{\pm 10.58}$ \\
& MUSE & $60.21_{\pm 8.72}$ & $86.46_{\pm 10.21}$ & $43.22_{\pm 13.31}$ & $78.30_{\pm 4.69}$ & $69.44_{\pm 10.20}$ & $\underline{86.75_{\pm 5.10}}$ & $66.87_{\pm 18.96}$ & $\underline{85.29_{\pm 4.38}}$ \\
\rowcolor{gray!15}& EviBall & $\mathbf{78.26_{\pm 11.24}}$ & $\mathbf{89.17_{\pm 4.17}}$ & $\mathbf{77.90_{\pm 14.33}}$ & $83.11_{\pm 8.39}$ & $\mathbf{88.34_{\pm 6.82}}$ & $\underline{87.30_{\pm 10.38}}$ & $\mathbf{86.56_{\pm 9.15}}$ & $\mathbf{85.41_{\pm 8.59}}$ \\
\midrule
\multirow{9}{*}{\rotatebox{90}{BRCA-Subtyping}}
& ABMIL & $57.70_{\pm 11.73}$ & $70.22_{\pm 13.97}$ & $57.59_{\pm 7.16}$ & $71.23_{\pm 9.24}$ & $60.13_{\pm 9.55}$ & $76.29_{\pm 10.52}$ & $74.34_{\pm 12.63}$ & $85.88_{\pm 7.91}$ \\
& CLAM & $55.73_{\pm 9.53}$ & $65.43_{\pm 14.73}$ & $51.99_{\pm 4.93}$ & $59.99_{\pm 5.74}$ & $60.01_{\pm 12.27}$ & $74.57_{\pm 9.17}$ & $75.82_{\pm 8.07}$ & $86.54_{\pm 7.01}$ \\
& TransMIL & $58.05_{\pm 3.22}$ & $66.30_{\pm 11.19}$ & $55.91_{\pm 8.36}$ & $64.86_{\pm 9.81}$ & $53.17_{\pm 3.89}$ & $64.97_{\pm 5.19}$ & $59.02_{\pm 7.01}$ & $78.37_{\pm 10.61}$ \\
& RRT & $58.46_{\pm 9.09}$ & $\underline{70.64_{\pm 11.38}}$ & $60.42_{\pm 6.59}$ & $74.50_{\pm 6.59}$ & $59.16_{\pm 9.29}$ & $\underline{81.54_{\pm 7.07}}$ & $\underline{77.13_{\pm 7.84}}$ & $\underline{87.58_{\pm 5.40}}$ \\
& DSMIL & $53.62_{\pm 5.75}$ & $57.93_{\pm 5.41}$ & $51.69_{\pm 2.70}$ & $59.15_{\pm 7.95}$ & $55.27_{\pm 4.85}$ & $66.41_{\pm 7.52}$ & $55.05_{\pm 6.52}$ & $67.14_{\pm 12.22}$ \\
& ViLaMIL & $59.22_{\pm 10.06}$ & $70.47_{\pm 14.30}$ & $\underline{65.79_{\pm 7.87}}$ & $\underline{76.68_{\pm 5.64}}$ & $\underline{61.61_{\pm 10.10}}$ & $75.00_{\pm 5.49}$ & $63.05_{\pm 13.20}$ & $70.37_{\pm 16.56}$ \\
& FOCUS & $\underline{59.25_{\pm 9.00}}$ & $66.91_{\pm 9.10}$ & $60.44_{\pm 8.58}$ & $70.51_{\pm 10.08}$ & $56.72_{\pm 5.51}$ & $72.20_{\pm 5.09}$ & $68.67_{\pm 11.31}$ & $81.36_{\pm 7.07}$ \\
& MUSE & $54.89_{\pm 7.63}$ & $64.26_{\pm 10.35}$ & $50.53_{\pm 1.75}$ & $60.69_{\pm 10.68}$ & $55.49_{\pm 4.12}$ & $64.19_{\pm 7.04}$ & $69.23_{\pm 4.55}$ & $78.19_{\pm 8.89}$ \\
\rowcolor{gray!15} & EviBall & $\mathbf{61.08_{\pm 8.25}}$ & $\mathbf{75.94_{\pm 8.25}}$ & $\mathbf{65.85_{\pm 6.22}}$ & $\mathbf{77.43_{\pm 3.31}}$ & $\mathbf{71.58_{\pm 12.67}}$ & $\mathbf{85.02_{\pm 6.57}}$ & $\mathbf{81.17_{\pm 6.93}}$ & $\mathbf{88.15_{\pm 5.94}}$ \\
\midrule
\multirow{9}{*}{\rotatebox{90}{IMP-Grading}}
& ABMIL & $45.22_{\pm 9.49}$ & $48.05_{\pm 10.98}$ & $\underline{72.27_{\pm 14.55}}$ & $\underline{73.40_{\pm 15.98}}$ & $\underline{78.85_{\pm 8.96}}$ & $\underline{80.32_{\pm 10.57}}$ & $83.33_{\pm 13.20}$ & $85.48_{\pm 8.23}$ \\
& CLAM & $43.24_{\pm 10.19}$ & $48.26_{\pm 8.30}$ & $59.63_{\pm 19.17}$ & $59.13_{\pm 16.15}$ & $75.75_{\pm 5.55}$ & $78.61_{\pm 5.76}$ & $\underline{85.59_{\pm 5.40}}$ & $\underline{87.08_{\pm 4.21}}$ \\
& TransMIL & $39.04_{\pm 5.27}$ & $46.84_{\pm 5.96}$ & $47.43_{\pm 5.98}$ & $54.66_{\pm 3.90}$ & $57.30_{\pm 7.42}$ & $63.08_{\pm 5.95}$ & $70.31_{\pm 13.63}$ & $73.26_{\pm 11.29}$ \\
& RRT & $\underline{47.44_{\pm 10.72}}$ & $49.74_{\pm 10.51}$ & $46.94_{\pm 18.09}$ & $48.00_{\pm 17.90}$ & $72.49_{\pm 13.51}$ & $78.47_{\pm 8.27}$ & $76.80_{\pm 18.37}$ & $77.49_{\pm 19.31}$ \\
& DSMIL & $43.36_{\pm 10.26}$ & $48.15_{\pm 15.56}$ & $53.53_{\pm 14.30}$ & $59.76_{\pm 12.07}$ & $71.47_{\pm 3.42}$ & $75.82_{\pm 5.02}$ & $81.29_{\pm 4.25}$ & $82.31_{\pm 3.87}$ \\
& ViLaMIL & $46.17_{\pm 10.15}$ & $40.30_{\pm 7.99}$ & $47.31_{\pm 10.93}$ & $45.31_{\pm 6.24}$ & $41.57_{\pm 9.92}$ & $45.07_{\pm 5.80}$ & $52.70_{\pm 5.16}$ & $51.69_{\pm 2.73}$ \\
& FOCUS & $46.06_{\pm 6.96}$ & $\underline{50.01_{\pm 8.24}}$ & $57.36_{\pm 6.90}$ & $61.65_{\pm 7.84}$ & $70.85_{\pm 8.32}$ & $73.06_{\pm 5.96}$ & $70.15_{\pm 12.46}$ & $68.88_{\pm 15.09}$ \\
& MUSE & $38.18_{\pm 6.26}$ & $41.48_{\pm 10.94}$ & $38.65_{\pm 2.48}$ & $46.14_{\pm 3.40}$ & $62.44_{\pm 18.45}$ & $65.87_{\pm 16.50}$ & $85.38_{\pm 5.07}$ & $84.46_{\pm 4.97}$ \\
\rowcolor{gray!15}& EviBall & $\mathbf{50.85_{\pm 9.91}}$ & $\mathbf{55.25_{\pm 11.35}}$ & $\mathbf{74.20_{\pm 7.73}}$ & $\mathbf{75.65_{\pm 4.41}}$ & $\mathbf{80.37_{\pm 4.82}}$ & $\mathbf{81.45_{\pm 4.02}}$ & $\mathbf{87.62_{\pm 2.93}}$ & $\mathbf{88.06_{\pm 3.62}}$ \\
\midrule
\multirow{10}{*}{\rotatebox{90}{BCNB-PR}}
& ABMIL & $51.90_{\pm 2.77}$ & $55.68_{\pm 6.72}$ & $50.57_{\pm 0.97}$ & $54.14_{\pm 7.52}$ & $53.22_{\pm 4.65}$ & $60.09_{\pm 7.22}$ & $50.71_{\pm 1.74}$ & $54.73_{\pm 7.76}$ \\
& CLAM & $\underline{56.15_{\pm 10.27}}$ & $57.41_{\pm 12.20}$ & $52.99_{\pm 4.81}$ & $55.42_{\pm 6.01}$ & $53.47_{\pm 4.82}$ & $60.64_{\pm 8.28}$ & $56.99_{\pm 1.63}$ & $62.10_{\pm 5.82}$ \\
& TransMIL & $52.07_{\pm 4.86}$ & $55.82_{\pm 7.56}$ & $49.11_{\pm 5.43}$ & $50.65_{\pm 7.68}$ & $54.63_{\pm 6.26}$ & $56.41_{\pm 10.67}$ & $50.28_{\pm 5.77}$ & $52.81_{\pm 6.65}$ \\
& RRT & $54.67_{\pm 5.51}$ & $56.06_{\pm 11.97}$ & $52.81_{\pm 4.35}$ & $54.60_{\pm 7.75}$ & $\underline{57.87_{\pm 9.46}}$ & $63.52_{\pm 8.74}$ & $56.09_{\pm 7.65}$ & $63.05_{\pm 9.21}$ \\
& DSMIL & $50.28_{\pm 0.94}$ & $48.49_{\pm 4.72}$ & $52.12_{\pm 2.36}$ & $53.10_{\pm 5.43}$ & $55.77_{\pm 8.79}$ & $58.66_{\pm 9.66}$ & $54.19_{\pm 6.78}$ & $55.41_{\pm 8.56}$ \\
& ViLaMIL & $54.98_{\pm 6.46}$ & $55.93_{\pm 11.18}$ & $\underline{55.89_{\pm 7.09}}$ & $55.38_{\pm 9.15}$ & $54.82_{\pm 7.42}$ & $61.86_{\pm 5.39}$ & $55.03_{\pm 5.07}$ & $63.55_{\pm 9.78}$ \\
& FOCUS & $50.60_{\pm 3.42}$ & $53.78_{\pm 9.48}$ & $47.94_{\pm 2.24}$ & $50.54_{\pm 8.51}$ & $56.15_{\pm 7.92}$ & $58.79_{\pm 8.27}$ & $53.57_{\pm 8.57}$ & $57.43_{\pm 11.42}$ \\
& MUSE & $50.98_{\pm 2.79}$ & $50.65_{\pm 1.97}$ & $50.21_{\pm 3.22}$ & $48.45_{\pm 7.61}$ & $52.34_{\pm 2.92}$ & $52.42_{\pm 8.22}$ & $51.30_{\pm 4.55}$ & $52.22_{\pm 12.06}$ \\
\rowcolor{gray!15}& EviBall-LLM & $55.05_{\pm 5.76}$ & $\mathbf{63.25_{\pm 8.48}}$ & $52.85_{\pm 4.04}$ & $\underline{58.48_{\pm 8.09}}$ & $57.47_{\pm 4.45}$ & $\underline{64.20_{\pm 7.10}}$ & $\mathbf{61.37_{\pm 6.50}}$ & $\underline{65.58_{\pm 5.23}}$ \\
\rowcolor{gray!15}& EviBall-RNA & $\mathbf{58.11_{\pm 6.41}}$ & $\underline{62.39_{\pm 8.01}}$ & $\mathbf{56.34_{\pm 8.73}}$ & $\mathbf{59.43_{\pm 9.22}}$ & $\mathbf{59.57_{\pm 8.15}}$ & $\mathbf{65.51_{\pm 9.99}}$ & $\underline{60.18_{\pm 2.80}}$ & $\mathbf{66.76_{\pm 4.32}}$ \\
\bottomrule
\end{tabular}
\caption{Few-shot performance on four WSI classification tasks. BCNB-PR additionally reports EviBall-RNA.}% Best and second-best results in each task-shot-metric block are shown in bold and underlined, respectively. BCNB-PR additionally reports EviBall-RNA.}
\label{tab:fewshot_selected_four_tasks}
\end{table*}
   %3.5 Training Objective and Inference
\subsection{Optimization}
\label{sec:optimization_inference}

EviBall is trained under the standard weakly supervised setting, where only slide-level labels are available. The class-wise logits produced by class-conditioned evidence retrieval are supervised with the standard cross-entropy loss $\mathcal{L}_{\mathrm{CE}}$ using the slide-level label.

To promote complementary evidence modeling within each local region, we introduce a same-region diversity regularizer for the two co-located evidence balls. For each valid region $\mathcal{R}_r$, let $c_{r_i}$ and $c_{r_j}$ denote the refined semantic centers of the two evidence balls initialized from the seed pair in Eq.~\eqref{init}. The diversity regularizer is defined as
\begin{equation}
\mathcal{L}_{\mathrm{div}}
=
\frac{1}{R_m}
\sum_{r=1}^{R_m}
\max\!\big(0,\ \mathrm{sim}(c_{r_i},c_{r_j})-\delta\big),
\end{equation}
where $R_m$ is the number of valid regions in the $m$-th WSI, $\mathrm{sim}(\cdot,\cdot)$ denotes cosine similarity, and $\delta$ is a similarity margin. This margin-based objective penalizes co-located evidence balls only when their semantic similarity exceeds $\delta$, thereby discouraging redundant local evidence representations.
The final training objective is $\mathcal{L}=(1 - \lambda_{\mathrm{div}})\mathcal{L}_{\mathrm{CE}}+\lambda_{\mathrm{div}}\mathcal{L}_{\mathrm{div}},$
where $\lambda_{\mathrm{div}}$ controls the strength of the diversity
regularization.

\section{Experiments}
\label{Experiments}
   %4.1 Datasets and Few-Shot Protocol
   %4.2 Baselines and Implementation Details
   %4.3 Main Results
   %4.4 Ablation Studies
   %4.5 Evidence Visualization and Analysis
   %4.6 Sensitivity and Robustness Analysis

\subsection{Experimental Settings}
\paragraph{Datasets.}
We evaluate EviBall on four few-shot WSI classification tasks built from five public datasets.
Specifically, RCC-Subtyping is constructed by combining CPTAC-CCRCC \cite{clark2019integrated} and DHMC-RCC \cite{zhu2021development}. BRCA-Subtyping is evaluated on TCGA-BRCA (Tomczak et al.~\citeyear{tomczak2015review}). IMP-Grading is evaluated on IMP-CRS 2024 \cite{oliveira2021cad}, and BCNB-PR prediction is evaluated on BCNB \cite{xu2021predicting}. 
%These tasks cover both morphology-oriented classification and molecular endpoint prediction. 
Detailed dataset descriptions and task definitions are provided in
Appendix.
%\footnote{https://portal.gdc.cancer.gov.}

\paragraph{Few-shot protocol.}
Following the common training-shot protocol, we perform patient-level five-fold cross-validation with a 6:2:2 train/validation/test split. Only the training pool is subsampled to $H$ WSIs per class ($H\in\{1,2,4,8\}$), while the validation and test folds remain unchanged. All methods use identical splits, nested support sets, validation data, and model-selection procedures.

\paragraph{Implementation details.}
We use AdamW as the optimizer with a learning rate of $2\times10^{-4}$.
Models are trained for up to 200 epochs with early stopping patience of 10. For fair comparison, all methods use the same frozen CONCH v1.5 patch encoder for visual feature extraction. For methods requiring textual class semantics, we use the TITAN text encoder as the common text encoder. All experiments are conducted on a single NVIDIA RTX 4090 GPU.

%\paragraph{Baselines.}
%We compare EviBall with representative WSI classification baselines, including attention-based MIL methods ABMIL (Ilse et al.~\citeyear{abmil}), CLAM~\cite{clam}, and DSMIL~\cite{li2021dual}, transformer-based MIL methods TransMIL~\cite{transmil} and RRT~\cite{rrt}, as well as recent structure-aware or few-shot WSI methods ViLaMIL~\cite{shi2024vila}, FOCUS~\cite{guo2025focus}, and MUSE \cite{Xu_2026_CVPR}. 

\paragraph{Evaluation metrics.}
Following the evaluation protocols of \cite{titan,zhang2026care}, we report balanced accuracy (ACC) as the primary metric to account for class imbalance. For binary classification tasks, AUC is used as the secondary metric, whereas F1 score is used for multi-class tasks with three or more classes. Results are reported as mean and standard deviation over five folds.
%\paragraph{Implementation details.} baseline 以及Evaluation Metrics.

\subsection{Main Results}

To validate the effectiveness of EviBall, we compare EviBall with representative WSI classification baselines, including attention-based MIL methods ABMIL (Ilse et al.~\citeyear{abmil}), CLAM~\cite{clam}, and DSMIL~\cite{li2021dual}, transformer-based MIL methods TransMIL~\cite{transmil} and RRT~\cite{rrt}, as well as recent structure-aware or few-shot WSI methods ViLaMIL~\cite{shi2024vila}, FOCUS~\cite{guo2025focus}, and MUSE \cite{Xu_2026_CVPR}. 
The results are reported in Table~\ref{tab:fewshot_selected_four_tasks}.
Overall, EviBall achieves the best performance in most task-shot-metric settings, demonstrating the effectiveness of structured evidence balls and class-conditioned evidence retrieval. Beyond the commonly evaluated multi-shot settings \cite{guo2025focus, Xu_2026_CVPR}, we additionally include the 1-shot regime as an extreme stress test. EviBall achieves the best or competitive performance across all four tasks. 
The performance gains remain consistent across the more stable multi-shot settings, indicating that EviBall is effective under varying levels of label scarcity. This consistency suggests that the improvements do not depend on a particular shot regime. Instead, class-conditioned retrieval over structured evidence balls provides more reliable evidence than global slide-level aggregation under limited supervision.

For the molecular endpoint task BCNB-PR, EviBall-RNA generally achieves stronger performance than EviBall-LLM, especially under 2/4/8-shot settings. This indicates that molecular-guided class queries provide more task-relevant semantics for molecular endpoint prediction than text-only class descriptions. Meanwhile, EviBall-LLM remains competitive in several settings, showing that the EviBall
framework can flexibly incorporate different semantic sources. These results support our design of using task-relevant class queries to retrieve and compete over evidence balls.
\subsection{Ablation Study}
\begin{figure*}[t]
\centering
\includegraphics[width=0.9\textwidth]{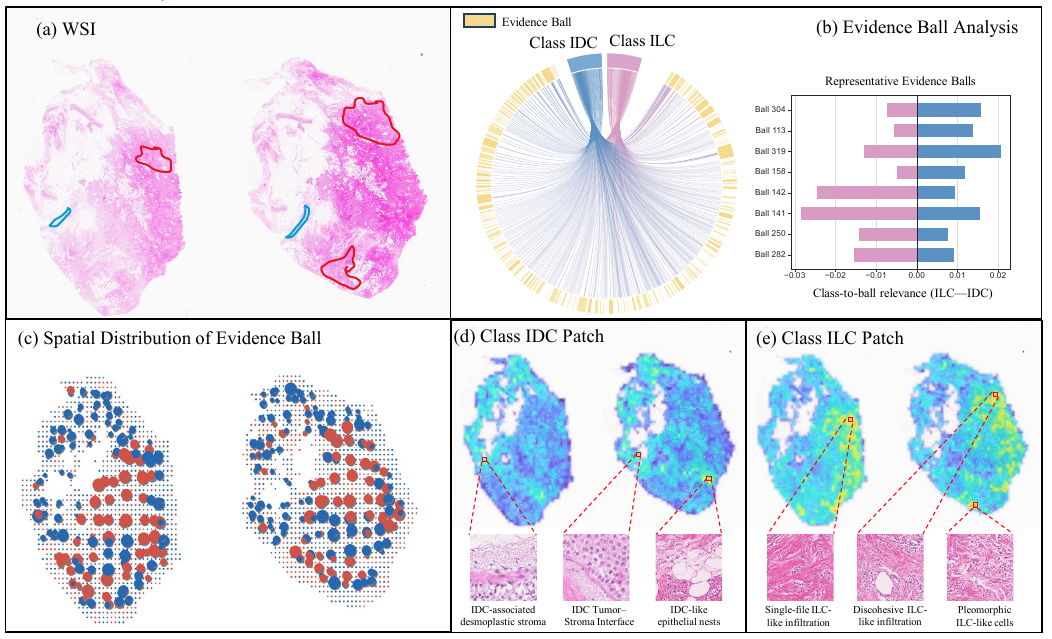} % Reduce the figure size so that it is slightly narrower than the column.
\caption{\textbf{Class-conditioned evidence analysis on a heterogeneous breast cancer slide.}
(a) H\&E whole-slide image.
(b) Differential relevance of representative evidence balls to IDC and ILC.
(c) Spatial distribution of class-associated evidence balls.
(d--e) IDC- and ILC-conditioned patch responses with representative histological regions.}
\label{fig:visualization}
\end{figure*}

To evaluate the contribution of each component, we conduct ablation studies on RCC-Subtyping under 1/2/4/8-shot settings, as shown in Table~\ref{tab:ablation}. Variant I replaces Evidence Balls with patch-level cross-attention. Variant II averages the class-conditioned representations into a shared slide feature. Variant III removes the spatial distance penalty in Eq.~\ref{LSSA}, while Variant IV removes the support mass bias from the support-aware cross-attention in Eq.~\ref{xatten}. Variant V removes the spatial-gated center update.

Variant I causes the largest performance drop, demonstrating the importance of organizing fragmented patches into structured evidence balls. Variant II also consistently degrades performance, particularly in low-shot settings, confirming the benefit of class-conditioned evidence retrieval over global feature aggregation. Removing the support bias or spatial-gated update generally reduces performance, showing that evidence support  mass and spatially aware refinement improve retrieval robustness. By comparison, removing spatial affinity has a smaller effect, suggesting that semantic similarity primarily determines patch-to-ball assignment, while spatial information mainly preserves evidence locality and coherence.
Overall, structured evidence ball construction and class-conditioned retrieval are the main contributors to EviBall, with support-aware and spatially aware components providing further improvements. To further assess robustness, the Appendix reports sensitivity analyses of $\lambda_{\mathrm{div}}$, the number of Evidence Balls $K$, and $\lambda_{\mathrm{spatial}}$, together with ablations on region-seed selection strategies and LLM choices for class-query generation.

\subsection{Interpretability and Evidence Analysis}
To examine how EviBall handles competing diagnostic evidence within a histologically heterogeneous slide, we analyze a breast cancer case originally diagnosed as mixed IDC--ILC but reclassified as ILC after centralized review of the TCGA slide, as shown in Fig.~\ref{fig:visualization}. After review by a pathologist, this WSI was found to contain limited evidence of IDC and extensive evidence of ILC.
EviBall retrieves evidence for both classes: the IDC query highlights desmoplastic stroma and epithelial nests, whereas the ILC query focuses on single-file infiltration, discohesive tumor cells, and pleomorphic ILC-like regions.
Although localized IDC-like cues remain, ILC-associated evidence is more spatially extensive and receives stronger collective class-to-ball relevance, leading to the final ILC prediction.

\begin{table}[t]
\centering
\small
\setlength{\tabcolsep}{3.5pt}
\begin{tabular}{lcccccccc}
\toprule
Method & \multicolumn{2}{c}{1-shot} & \multicolumn{2}{c}{2-shot} & \multicolumn{2}{c}{4-shot} & \multicolumn{2}{c}{8-shot} \\
\cmidrule(lr){2-3}\cmidrule(lr){4-5}\cmidrule(lr){6-7}\cmidrule(lr){8-9}
& ACC & F1 & ACC & F1 & ACC & F1 & ACC & F1 \\
\midrule
I & 75.7 & 76.9 & 75.1 & 75.6 & 76.0 & 78.3 & 84.3 & 81.0 \\
II & 69.2 & 82.3 & 68.8 & 82.4 & 79.2 & 86.3 & 82.8 & 78.4 \\
III & 77.2 & 88.6 & 77.3 & 82.0 & 88.3 & 87.2 & $\bm{86.9}$ & 83.5 \\
IV & 76.3 & 84.9 & 76.9 & 78.9 & 87.7 & 84.5 & 85.0 & 85.2 \\
V & 77.9 & 86.6 & 77.6 & 82.4 & 86.7 & 83.9 & 83.0 & 82.6 \\
\rowcolor{gray!15} EviBall & $\bm{78.3}$ & $\bm{89.2}$ & $\bm{77.9}$ & $\bm{83.1}$ & $\bm{88.3}$ & $\bm{87.3}$ &${86.6}$ & $\bm{85.4}$ \\
\bottomrule
\end{tabular}
\caption{Ablation study on RCC-Subtyping. Variant definitions are described in the text.}%Variant I bypasses Evidence Ball generation and directly performs patch-level cross-attention, Variant II removes class-wise evidence prediction by averaging class-conditioned evidence representations into a shared slide-level feature, followed by a linear classifier. Variant III removes spatial information from patch-to-ball assignment, Variant IV removes the quality bias in evidence decoding, Variant V removes the spatial-gated center update.}
\label{tab:ablation}
\end{table}
\section{Conclusion}

We presented EviBall, a class-conditioned evidence retrieval framework for few-shot WSI classification. EviBall organized local patches into structured evidence balls and retrieved them using task-relevant class queries, enabling class-specific reasoning beyond global slide aggregation. Experiments across morphology-oriented and molecular endpoint tasks demonstrated consistent improvements over existing baselines, while providing spatially localized and class-specific evidence for interpretable prediction.

%we propose CARE, a slide-level foundation model that tackles this problem through an adaptive region generator. In addition, we design an efficient pretraining pipeline that achieves competitive performance while using substantially less data. Experimental results demonstrate that CARE surpasses current slide-level foundation models and establishes an efficient, reliable route for computational pathology. Its architecture and pretraining provide a pathology-grounded, efficient route to diagnostically faithful modeling.
\section*{Acknowledgements}
This work was partially supported by the National Natural Science Foundation of China (Grant No. 62506291), the National Science and Technology Major Project (Grant No. 2025ZD0544802), the Key Research and Development Program of Shaanxi Province (Grant No. 2024SF-GJHX-32), the Key Research and Development Program of Ningxia Hui Autonomous Region (Grant No. 2023BEG02023), the Noncommunicable Chronic Diseases–National Science and Technology Major Project (Grant No. 2024ZD0527700), the project ``Research on Key Technologies for Full-Chain Intelligent Pathological Diagnosis'' (Grant No. HX202440) of The First Affiliated Hospital of Xi'an Jiaotong University, and GE HealthCare.
{
    \small
    \bibliographystyle{ieeenat_fullname}
    \bibliography{main}

@String(CVPR= {IEEE Conf. Comput. Vis. Pattern Recog.})

@String(CVPR  = {CVPR})

@inproceedings{zhang2026care,
  title={{CARE}: A Molecular-Guided Foundation Model with Adaptive Region Modeling for Whole Slide Image Analysis},
  author={Zhang, Di and Gong, Zhangpeng and Pang, Xiaobo and Liu, Jiashuai and Lu, Junbo and Cui, Hao and Ge, Jiusong and Zeng, Zhi and Yi, Kai and Li, Yinghua and others},
  booktitle={Proceedings of the IEEE/CVF Conference on Computer Vision and Pattern Recognition},
  pages={21078--21088},
  year={2026}
}

@article{zheng2022graph,
  title={A graph-transformer for whole slide image classification},
  author={Zheng, Yi and Gindra, Rushin H and Green, Emily J and Burks, Eric J and Betke, Margrit and Beane, Jennifer E and Kolachalama, Vijaya B},
  journal={IEEE transactions on medical imaging},
  volume={41},
  number={11},
  pages={3003--3015},
  year={2022},
  publisher={IEEE}
}

@article{zhang2025stadis,
  title={{StaDis}: Stability distance to detecting out-of-distribution data in computational pathology},
  author={Zhang, Di and Ge, Jiusong and Liu, Jiashuai and Wang, Chunbao and Gong, Tieliang and Gao, Zeyu and Li, Chen},
  journal={Medical Image Analysis},
  pages={103774},
  year={2025},
  publisher={Elsevier}
}

@article{wong2026few,
  title={Few-shot learning from gigapixel images via hierarchical vision-language alignment and modeling},
  author={Wong, Bryan and Kim, Jongwoo and Fu, Huazhu and Yi, Mun},
  journal={Advances in Neural Information Processing Systems},
  volume={38},
  pages={71313--71354},
  year={2026}
}

@article{zhu2021development,
title={Development and evaluation of a deep neural network for histologic classification of renal cell carcinoma on biopsy and surgical resection slides},
author={Zhu, Mengdan and Ren, Bing and Richards, Ryland and Suriawinata, Matthew and Tomita, Naofumi and Hassanpour, Saeed},
journal={Scientific reports},
volume={11},
number={1},
pages={1--9},
year={2021},
publisher={Nature Publishing Group}
}

@article{weishaupt2025evidence,
  title={Evidence-based diagnostic reasoning with multi-agent copilot for human pathology},
  author={Weishaupt, Luca L and Chen, Chengkuan and Williamson, Drew FK and Chen, Richard J and Jaume, Guillaume and Ding, Tong and Chen, Bowen and Vaidya, Anurag and Le, Long Phi and Lu, Ming Y and others},
  journal={arXiv preprint arXiv:2506.20964},
  year={2025}
}

@article{liu2024correlation,
  title={From Correlation to Causation: Max-Pooling-Based Multi-Instance Learning Leads to More Robust Whole Slide Image Classification},
  author={Liu, Xin and Zhang, Weijia and Tang, Wei and Le, Thuc Duy and Li, Jiuyong and Liu, Lin and Zhang, Min-Ling},
  journal={arXiv preprint arXiv:2408.09449},
  year={2024}
}

@article{xu2021predicting,
	title={Predicting axillary lymph node metastasis in early breast cancer using deep learning on primary tumor biopsy slides},
	author={Xu, Feng and Zhu, Chuang and Tang, Wenqi and Wang, Ying and Zhang, Yu and Li, Jie and Jiang, Hongchuan and Shi, Zhongyue and Liu, Jun and Jin, Mulan},
	journal={Frontiers in Oncology},
	volume={11},
	pages={759007},
	year={2021},
	publisher={Frontiers Media SA}
}

@article{tomczak2015review,
  title={Review The Cancer Genome Atlas (TCGA): an immeasurable source of knowledge},
  author={Tomczak, Katarzyna and Czerwi{\'n}ska, Patrycja and Wiznerowicz, Maciej},
  journal={Contemporary Oncology/Wsp{\'o}{\l}czesna Onkologia},
  volume={2015},
  number={1},
  pages={68--77},
  year={2015},
  publisher={Termedia}
}

@article{clark2019integrated,
  title={Integrated proteogenomic characterization of clear cell renal cell carcinoma},
  author={Clark, David J and Dhanasekaran, Saravana M and Petralia, Francesca and Pan, Jianbo and Song, Xiaoyu and Hu, Yingwei and da Veiga Leprevost, Felipe and Reva, Boris and Lih, Tung-Shing M and Chang, Hui-Yin and others},
  journal={Cell},
  volume={179},
  number={4},
  pages={964--983},
  year={2019},
  publisher={Elsevier}
}

@article{oliveira2021cad,
	title={{CAD} systems for colorectal cancer from WSI are still not ready for clinical acceptance},
	author={Oliveira, Sara P and Neto, Pedro C and Fraga, Joao and Montezuma, Diana and Monteiro, Ana and Monteiro, Jo{\~a}o and Ribeiro, Liliana and Gon{\c{c}}alves, Sofia and Pinto, Isabel M and Cardoso, Jaime S},
	journal={Scientific Reports},
	volume={11},
	number={1},
	pages={14358},
	year={2021},
	publisher={Nature Publishing Group UK London}
}

@article{clam,
	title={Data-efficient and weakly supervised computational pathology on whole-slide images},
	author={Lu, Ming Y and Williamson, Drew FK and Chen, Tiffany Y and Chen, Richard J and Barbieri, Matteo and Mahmood, Faisal},
	journal={Nature Biomedical Engineering},
	volume={5},
	number={6},
	pages={555--570},
	year={2021},
	publisher={Nature Publishing Group UK London}
}

@inproceedings{li2021dual,
	title={Dual-stream multiple instance learning network for whole slide image classification with self-supervised contrastive learning},
	author={Li, Bin and Li, Yin and Eliceiri, Kevin W},
	booktitle={Proceedings of the IEEE/CVF conference on Computer Vision and Pattern Recognition},
	pages={14318--14328},
	year={2021}
}

@inproceedings{li2026turning,
  title={Turning Pre-Trained Vision Transformers into End-to-End Histopathology Whole Slide Image Models for Survival Prediction},
  author={Li, Jiawen and Hu, Jiali and Ling, Xitong and Yan, Renao and Chen, Yuxuan and Guan, Tian and He, Yonghong},
  booktitle={Proceedings of the IEEE/CVF Conference on Computer Vision and Pattern Recognition},
  pages={21046--21056},
  year={2026}
}

@InProceedings{Xu_2026_CVPR,
    author    = {Xu, Jiahao and Huang, Sheng and Zhang, Xin and Nan, Zhixiong and Dong, Jiajun and Mu, Nankun},
    title     = {MUSE: Harnessing Precise and Diverse Semantics for Few-Shot Whole Slide Image Classification},
    booktitle = {Proceedings of the IEEE/CVF Conference on Computer Vision and Pattern Recognition (CVPR)},
    month     = {June},
    year      = {2026},
    pages     = {33911-33921}
}

@article{yao2020whole,
	title={Whole slide images based cancer survival prediction using attention guided deep multiple instance learning networks},
	author={Yao, Jiawen and Zhu, Xinliang and Jonnagaddala, Jitendra and Hawkins, Nicholas and Huang, Junzhou},
	journal={Medical image analysis},
	volume={65},
	pages={101789},
	year={2020},
	publisher={Elsevier}
}

@inproceedings{li2025similarity,
  title={Similarity-Aware Dual-Perspective Learning for Medical Event Prediction},
  author={Li, Yang and Ge, Jiusong and Cao, Shilei and Zhan, Yingkang and Gong, Zhangpeng and Niu, Jiawei and Liu, Jiashuai and Zhang, Di and Li, Chen},
  booktitle={2025 IEEE International Conference on Bioinformatics and Biomedicine (BIBM)},
  pages={6263--6270},
  year={2025},
  organization={IEEE}
}

@article{gao2025alpaca,
  title={ALPaCA: Adapting Llama for Pathology Context Analysis to enable slide-level question answering},
  author={Gao, Zeyu and He, Kai and Su, Weiheng and Machado, Ines P and McGough, William and Jimenez-Linan, Mercedes and Rous, Brian and Wang, Chunbao and Li, Chengzu and Pang, Xiaobo and others},
  journal={medRxiv},
  pages={2025--04},
  year={2025},
  publisher={Cold Spring Harbor Laboratory Press}
}

@article{huang2023visual,
	title={A visual--language foundation model for pathology image analysis using medical twitter},
	author={Huang, Zhi and Bianchi, Federico and Yuksekgonul, Mert and Montine, Thomas J and Zou, James},
	journal={Nature Medicine},
	volume={29},
	number={9},
	pages={2307--2316},
	year={2023},
	publisher={Nature Publishing Group US New York}
}

@article{song2023artificial,
	title={Artificial intelligence for digital and computational pathology},
	author={Song, Andrew H and Jaume, Guillaume and Williamson, Drew FK and Lu, Ming Y and Vaidya, Anurag and Miller, Tiffany R and Mahmood, Faisal},
	journal={Nature Reviews Bioengineering},
	volume={1},
	number={12},
	pages={930--949},
	year={2023},
	publisher={Nature Publishing Group UK London}
}

@inproceedings{niu2025learning,
  title={Learning Heterogeneous Embedding with Prototype-Aware Graph Attention for Whole Slide Image Classification},
  author={Niu, Yi and Liu, Jiashuai and Zhan, Yingkang and Shi, Jiangbo and Chen, Jian and Zhang, Di and Li, Chen and Gao, Zeyu},
  booktitle={2025 IEEE International Conference on Bioinformatics and Biomedicine (BIBM)},
  pages={2671--2678},
  year={2025},
  organization={IEEE}
}

@article{niu2026age,
  title={AGE-MIL: Anchor-Guided Evidence Learning for Patient-Level Prediction},
  author={Niu, Jiawei and Chen, Jian and Zhang, Di and Lu, Junbo and Liao, Zhangcheng and Liu, Xuhao and Zhong, Honglin and Crispin-Ortuzar, Mireia and Li, Chen and Gao, Zeyu and others},
  journal={arXiv preprint arXiv:2606.12126},
  year={2026}
}

@article{yan2026llm,
  title={LLM-Guided Diagnostic Evidence Alignment for Medical Vision-Language Pretraining under Limited Pairing},
  author={Yan, Huimin and Bai, Liang and Yang, Xian and Chen, Long},
  journal={arXiv preprint arXiv:2602.07540},
  year={2026}
}

@inproceedings{shi2024vila,
	title={Vila-mil: Dual-scale vision-language multiple instance learning for whole slide image classification},
	author={Shi, Jiangbo and Li, Chen and Gong, Tieliang and Zheng, Yefeng and Fu, Huazhu},
	booktitle={Proceedings of the IEEE/CVF Conference on Computer Vision and Pattern Recognition},
	pages={11248--11258},
	year={2024}
}

@article{conch,
  title={A visual-language foundation model for computational pathology},
  author={Lu, Ming Y and Chen, Bowen and Williamson, Drew FK and Chen, Richard J and Liang, Ivy and Ding, Tong and Jaume, Guillaume and Odintsov, Igor and Le, Long Phi and Gerber, Georg and others},
  journal={Nature Medicine},
  volume={30},
  number={3},
  pages={863--874},
  year={2024},
  publisher={Nature Publishing Group US New York}
}

@inproceedings{abmil,
	title={Attention-based deep multiple instance learning},
	author={Ilse, Maximilian and Tomczak, Jakub and Welling, Max},
	booktitle={Proceedings of the International Conference on Machine Learning},
	pages={2127--2136},
	year={2018},
}

@article{cui2023scGPT,
  title={scGPT: toward building a foundation model for single-cell multi-omics using generative AI},
  author={Cui, Haotian and Wang, Chloe and Maan, Hassaan and Pang, Kuan and Luo, Fengning and Duan, Nan and Wang, Bo},
  journal={Nature methods},
  volume={21},
  number={8},
  pages={1470--1480},
  year={2024},
  publisher={Nature Publishing Group US New York}
}

@inproceedings{rrt,
	title={Feature re-embedding: Towards foundation model-level performance in computational pathology},
	author={Tang, Wenhao and Zhou, Fengtao and Huang, Sheng and Zhu, Xiang and Zhang, Yi and Liu, Bo},
	booktitle={Proceedings of the IEEE/CVF Conference on Computer Vision and Pattern Recognition},
	pages={11343--11352},
	year={2024}
}

@article{transmil,
	title={Trans{MIL}: Transformer based correlated multiple instance learning for whole slide image classification},
	author={Shao, Zhuchen and Bian, Hao and Chen, Yang and Wang, Yifeng and Zhang, Jian and Ji, Xiangyang and others},
	journal={Advances in Neural Information Processing Systems},
	volume={34},
	pages={2136--2147},
	year={2021}
}

@article{titan,
	title={A multimodal whole-slide foundation model for pathology},
	author={Ding, Tong and Wagner, Sophia J and Song, Andrew H and Chen, Richard J and Lu, Ming Y and Zhang, Andrew and Vaidya, Anurag J and Jaume, Guillaume and Shaban, Muhammad and Kim, Ahrong and others},
	journal={Nature Medicine},
	volume={31},
	number={11},
	pages={3749--3761},
	year={2025}
}

@inproceedings{guo2025focus,
  title={{Focus}: Knowledge-enhanced adaptive visual compression for few-shot whole slide image classification},
  author={Guo, Zhengrui and Xiong, Conghao and Ma, Jiabo and Sun, Qichen and Feng, Lishuang and Wang, Jinzhuo and Chen, Hao},
  booktitle={Proceedings of the Computer Vision and Pattern Recognition Conference},
  pages={15590--15600},
  year={2025}
}

@article{fu2010milis,
  title={MILIS: Multiple instance learning with instance selection},
  author={Fu, Zhouyu and Robles-Kelly, Antonio and Zhou, Jun},
  journal={IEEE Transactions on Pattern Analysis and Machine Intelligence},
  volume={33},
  number={5},
  pages={958--977},
  year={2010},
  publisher={IEEE}
}

@inproceedings{li2026universal,
  title={Universal-to-Specific: Dynamic Knowledge-Guided Multiple Instance Learning for Few-Shot Whole Slide Image Classification},
  author={Li, Junjian and Kuang, Hulin and Liu, Jin and Yue, Hailin and He, Mengshen and Wang, Jianxin},
  booktitle={Proceedings of the IEEE/CVF Conference on Computer Vision and Pattern Recognition},
  pages={26614--26623},
  year={2026}
}
}


\end{document}